# Multi-Scale Prototypical Transformer for Whole Slide Image Classification

Saisai Ding[1], Jun Wang[1], Juncheng Li[1], and Jun Shi[1✉]

[1] School of Communication and Information Engineering, Shanghai University, Shanghai, China
junshi@shu.edu.cn

**Abstract.** Whole slide image (WSI) classification is an essential task in computational pathology. Despite the recent advances in multiple instance learning (MIL) for WSI classification, accurate classification of WSIs remains challenging due to the extreme imbalance between the positive and negative instances in bags, and the complicated pre-processing to fuse multi-scale information of WSI. To this end, we propose a novel multi-scale prototypical Transformer (MSPT) for WSI classification, which includes a prototypical Transformer (PT) module and a multi-scale feature fusion module (MFFM). The PT is developed to reduce redundant instances in bags by integrating prototypical learning into the Transformer architecture. It substitutes all instances with cluster prototypes, which are then re-calibrated through the self-attention mechanism of Transformer. Thereafter, an MFFM is proposed to fuse the clustered prototypes of different scales, which employs MLP-Mixer to enhance the information communication between prototypes. The experimental results on two public WSI datasets demonstrate that the pro-posed MSPT outperforms all the compared algorithms, suggesting its potential applications.

## 1 Introduction

Histopathological images are regarded as the 'gold standard' in the diagnosis of cancers. With the advent of the whole slide image (WSI) scanner, deep learning has gained its reputation in the field of computational pathology [1, 2, 3]. However, WSIs are extremely large in the size and lack of pixel-level annotations, making it difficult to adopt the traditional supervised learning methods for WSI classification [4].

To address this issue, multiple instance learning (MIL) has been successfully applied to the WSI classification task as a weakly supervised learning problem [5, 6, 7]. In this context, a WSI is considered as a bag, and the cropped patches within the slide are the instances in this bag. However, the lesion regions usually only account for a small portion of the WSI, resulting in a large number of negative patches. When the positive and negative instances in the bag are highly imbalanced, the MIL models are prone to in-correctly discriminate these positive instances when using simple aggregation operations. To this end, several attention-based MIL models, such as ABMIL [8] and DSMIL [9], apply variants of the attention mechanism to re-weight instance features. Thereafter, the recent works develop the Transformer-based architectures to better model long-range instance correlations via self-attention [10, 11, 12, 13]. However, since the average bag size of a WSI is more than 8000 at 20× magnification, it is computationally infeasible to use the conventional Transformer and other stacked self-attention network architectures in MIL-related tasks.

Recently, prototypical learning is applied in WSI analysis to identify representative instances in the bag [14]. Some works adopt the $K$-means clustering on all instances in a bag to obtain $K$ cluster centers i.e., instance prototypes, and then use these prototypes to represent the bags [15, 16]. These clustering-based MIL algorithms can significantly reduce the redundant instances, and thereby improving the training efficiency for WSI classification. However, it is different for $K$-means to specify the cluster number as well as the initial cluster centers, and different initial values may lead to different cluster results, thus affecting the performance of MIL. Besides, affected by the feature extractor, the clustering-based MIL algorithms may ignore the most important instances that contain critical diagnostic information. Therefore, it is necessary to develop a method that can fully exploit the potential complementary information between critical instances and prototypes to improve representation learning of prototypes.

On the other hand, when pathologists analysis the WSIs, they always observe the tissues at various resolutions [17]. Inspired by this diagnostic manner, some works use multi-scale information of WSIs to improve diagnostic accuracy. For example, Li et al. [9] adopted a pyramidal concatenation mechanism to fuse the multi-scale features of WSIs, in which the feature vectors of low-resolution patches are replicated and concatenated with the those of their corresponding high-resolution patches; Hou et al. [18] propose a heterogeneous graph neural network to learn the hierarchical representation of WSIs from a heterogeneous graph, which is constructed by the feature and spatial-scaling relationship of multi-resolution patches. However, since the number of patches at each resolution is quite different, it requires complex pre-processing to spatially align feature vectors of patches in different resolutions. Therefore, it is significant to develop an efficient and effective patch aggregation strategy to learn multi-scale information from WSIs.

In this work, we propose a Multi-Scale Prototypical Transformer (MSPT) for WSI classification. The MSPT includes two key components: a prototypical Transformer (PT) and a multi-scale feature fusion module (MFFM). The specifically developed PT uses a clustering algorithm to extract instance prototypes from the bags, and then re-calibrates these prototypes at each scale with the self-attention mechanism in Transformer [19]. MFFM is designed to effectively fuse multi-scale information of WSIs, which utilizes the MLP-Mixer [20] to learn effective representations by aggregating the multi-scale prototypes generated by the PT. The MLP-Mixer adopts two types of MLP layers to allow information communication in different dimensions of data.

The contributions of this work are summarized as follows:

1) A novel prototypical Transformer (PT) is proposed to learn superior prototype representation for WSI classification by integrating prototypical learning into the Transformer architecture. It can effectively re-calibrate the cluster prototypes as well as reduce the computational complexity of the Transformer.

2) A new multi-scale feature fusion module (MFFM) is developed based on the MLP-Mixer to enhance the information communication among phenotypes. It can effectively capture multi-scale information in WSI to improve the performance of WSI classification.

## 2 Method

### 2.1 MIL Problem Formulation

MIL is a typical weakly supervised learning method, where the training data consists of a set of bags, and each bag contains multiple instances. The goal of MIL is to learn a classifier that can predict the label of a bag based on the instances in it. In binary classification, a bag can be marked as negative if all in-stances in the bag are negative, otherwise, the bag is labeled as positive with at least one positive instance. In the MIL setting, a WSI is considered as a bag and the numerous cropped patches in WSI are regarded as instances in the bag. A WSI dataset T can be defined as:

$$T = \{x_i, y_i\}_{i=1}^{i=N}, x_i = \{I_i^j\}_{j=1}^{j=n} \tag{1}$$

where $x_i$ denotes a patient, $y_i$ the label of $x_i$, $I_i^j$ is the $j$-th instance of $x_i$, $N$ is the number of patients and $n$ is the number of instances.

### 2.2 Multi-scale Prototypical Transformer (MSPT)

The overall architecture of MSPT is shown in Fig. 1. A WSI is first divided into non-overlapping patches at different resolutions, and a pre-trained ResNet18 [21] is used to extract features from each patch. The learned multi-scale features are then fed into the proposed MSPT, which consists of a PT and an MFFM, to re-calibrate cluster prototypes at each scale and fuse multi-scale information of WSI. Finally, a WSI-level classifier is trained to predict the bag label.

**Pre-training.** It is a time consuming and tedious task for pathologists to annotate the patch-level labels in gigapixel WSIs, thus, a common practice is to use a pre-trained encoder network to extract instance-level features, such as an ImageNet pre-trained encoder or a self-supervised pre-trained encoder. In this work, we follow [9] to adopt SimCLR [22] to pre-training the patch encoder at different resolutions. SimCLR is a self-supervised learning algorithm to pre-trainng a network by maximizing the similarity between positive pairs and minimizing the similarity between negative pairs [22]. After pre-training, the extracted instances of different scales are fed into MSPT for prototype learning and multi-scale learning.

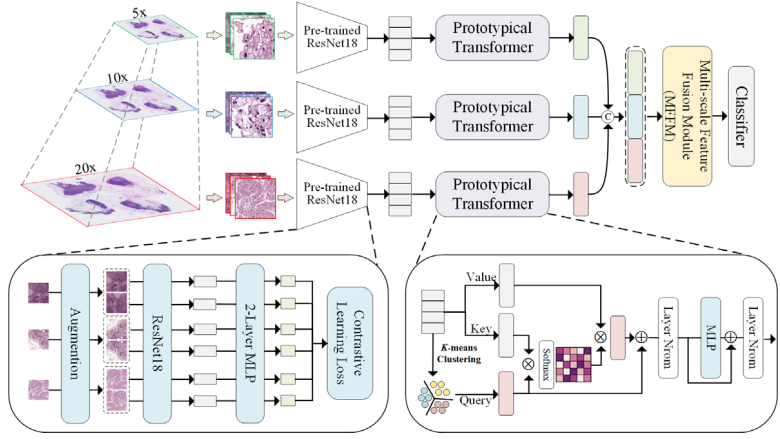

**Fig. 1.** Overview of the proposed MSPT.

**Prototypical Transformer (PT).** Most tissues in WSIs are redundancy, and therefore, we introduce the instance prototypes to reduce redundant instances. Specifically, for each instance bag $X_{bag} \in \mathbb{R}^{n \times d_k}$, the $K$-means clustering algorithm is applied on all instances to get $K$ centers (prototypes). These cluster prototypes can be used as instances to represent a new bag $P_{bag} \in \mathbb{R}^{k \times d_k}$. However, the $K$-means clustering algorithm is sensitive to the initial selection of cluster centers, i.e. different initializations can lead to different results, and the final result may not be the global optimal solution. It is essential to try different initializations and choose the one with the lowest error. However, the WSI dataset generally has a long sequence of instances, which makes the clustering algorithms computationally expensive and slow down as the size of the bag increases.

To solve the issue above, we propose to apply the self-attention (SA) mechanism in Transformer to re-calibrate these cluster prototypes. As shown in Fig.1, the optimization process can be divided into two steps: 1) the initial cluster prototype bag is obtained in the pre-processing stage by using the $K$-means clustering on $X_{bag}$; 2) PT uses $X_{bag}$ to optimize $P_{bag}$ via the self-attention mechanism in Transformer. The detailed process is as follows:

$$\text{SA}(P_{bag}, X_{bag}) = softmax\left(\frac{QK^T}{\sqrt{d_k}}\right) \cdot V$$

$$= softmax\left(\frac{W_q P_{bag}(X_{bag}W_k)^T}{\sqrt{d_k}}\right) W_v X_{bag} \rightarrow A_{map} W_v X_{bag} \rightarrow \hat{P} \quad (2)$$

where $W_q$, $W_k$, $W_v \in \mathbb{R}^{d_k \times d_k}$ are trainable matrices of query $P_{bag}$ and the key-value pair $(X_{bag}, X_{bag})$, respectively, and $A_{map} \in \mathbb{R}^{k \times n}$ is the attention matrix to compute the weight of $X_{bag}$. Thus, the computational complexity of SA is $O(nm)$ instead of $O(n^2)$, and the $k$ is much less than $n$. Specifically, for a single clustering prototype $p_k \in P$, the SA layer scores the pairwise similarity between $p_k$ and $x_n$ for all $x_n \in X$, which can be written as a row vector $[a_{k1}, a_{k2}, a_{k3}, ..., a_{kn}]$ in $\boldsymbol{A}_{map}$. These attention scores are then weighted to $\boldsymbol{X}_{bag}$ to update the $p_k \in \mathbb{R}^{1 \times d_k}$ for completing the calibration of the clustering prototypes $\hat{\boldsymbol{P}} \in \mathbb{R}^{k \times d_k}$.

As mentioned above, existing clustering-based MIL methods use the $K$-means clustering to identify instances prototypes in the bag, where the most important instances that contain the key semantic information may be ignored. On the contrary, our PT can efficiently use all the instances to update the cluster prototypes multiple times. Therefore, the combination of bag instances is no longer static and fixed, but diverse and dynamic. It means that different new bags can be fed into the MFFM each time. In addition, by applying PT to each scale, the number of cluster prototypes obtained at different scales is consistent, so there is no need for additional operations to align multi-scale features.

**Multi-scale Feature Fusion Module (MFFM).** To fuse the output clustered prototypes at different scales in MSPT, we proposed an MFFM, which consists of an MLP-Mixer and a Gated Attention Pooling (GAP). The MLP-Mixer is used to enhance the information communication of the prototype representation, and the GAP is used to get the WSI-level representation for WSI classification.

As shown in Fig. 2, The Mixer layer of MLP-Mixer contains one token-mixing MLP and one channel-mixing MLP, each consisting of two fully-connected layers and a GELU activation function [23]. Token-mixing MLP is a cross-location operation to mix all prototypes, while channel-mixing MLP is a pre-location operation to mix features of each prototype. Thus, MLP-Mixer allows the information communication between different prototypes and prototype features to learn superior representation through information aggregation.

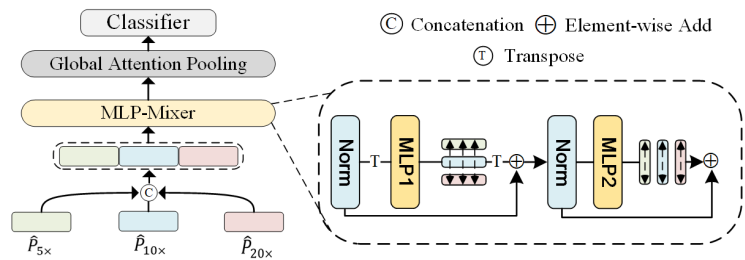

**Fig. 2.** The structure of MFFM.

Specifically, the procedure of MFFM is described as follows:

We first perform the feature concatenation operation on the multi-scale output clustering prototypes $[\hat{P}_{20\times}, \hat{P}_{10\times}, \hat{P}_{5\times}]$ to construct a feature pyramid $\bar{P}$:

$$concat[\hat{P}_{20\times}, \hat{P}_{10\times}, \hat{P}_{5\times}] \rightarrow \bar{P} \in \mathbb{R}^{k \times 3d_k} \tag{3}$$

where $d_k$ is the feature vector dimension of the prototypes.

Then, the $\breve{P}$ is fed to the MLP-Mixer to obtain the corresponding hidden feature representation $H \in \mathbb{R}^{k \times 3d_k}$ as follows:

$$H_1 = \breve{P}^T + W_2\sigma(W_1\text{LN}(\breve{P}^T))$$
$$H = H_1^T + W_4\sigma(W_3\text{LN}(H_1^T)) \tag{4}$$

where LN denotes the layer normalization, $\sigma$ denotes the activation function implemented by GELU, $W_1 \in \mathbb{R}^{k \times c}, W_2 \in \mathbb{R}^{c \times k}, W_3 \in \mathbb{R}^{3d_k \times d_s}$ and $W_4 \in \mathbb{R}^{d_s \times 3d_k}$ are the weight matrices of MLP layers. $c$ and $d_s$ are tunable hidden widths in the token-mixing and channel-mixing MLP, respectively.

Finally, the $H$ is fed to the gated attention pooling (GAP) [8] to get the WSI-level representation $Z \in \mathbb{R}^{1 \times 3d_k}$ for WSI classification:

$$Z = GAP(H)$$
$$\hat{Y} = softmax(MLP(Z)) \tag{5}$$

where $\hat{Y} \in \mathbb{R}^{1 \times d_{out}}$ is the class label probability of the bag, and $d_{out}$ is the number of classes.

## 3 Experiments and Results

### 3.1 Datasets

To evaluate the effectiveness of MSPT, we conducted experiments on two public dataset, namely Camelyon16 [24] and TCGA-NSCLC. Camelyon16 is a WSI dataset for the automated detection of metastases in lymph node tissue slides. It includes 270 training samples and 129 testing samples. After pre-processing, a total of 2.4 million patches at ×20 magnification, 0.56 million patches at ×10 magnification, and 0.16 million patches at ×5 magnification, with an average of about 5900, 1400, and 400 patches per bag. The TCGA-NSCLC dataset includes two sub-types of lung cancer, i.e., Lung Squamous Cell Carcinoma (TGCA-LUSC) and Lung Adenocarcinoma (TCGA-LUAD). We collected a total of 854 diagnostic slides from the National Cancer Institute Data Portal (https://portal.gdc.cancer.gov). The dataset yields 4.3 million patches at 20× magnification, 1.1 million patches at 10× magnification, and 0.30 million patches at 5× magnification with an average of about 5000, 1200, and 350 patches per bag.

### 3.2 Experiment Setup and Evaluation Metrics.

In WSI pre-processing, each slide is cropped into non-overlapping $256 \times 256$ patches at different magnifications, and a threshold is set to filter out background ones. After patching, we use a pre-trained ResNet18 model to convert each $256 \times 256$ patch into a 512- dimensional feature vector. We selected accuracy (ACC) and area under curve (AUC) as evaluation metrics. For Camelyon16 dataset, we reported the results of the official testing set. For TCGA-NSCLC, we conducted five cross-validation on the 854 slides, and the results are reported in the format of mean ± SD (standard deviation).

### 3.3 Implementation Details.

For the feature extractor, we employed the SimCLR encoder trained by Lee et al. [9] for the Camelyon16 and TCGA datasets. But [9] only trained SimCLR encoders at 20× and 5× magnification, to align with that setting, we used the same settings to train the SimCLR encoder at 10× magnification on both datasets. For the proposed MSPT, the Adam optimizer was used to update the model weights, the initial learning rate of 1e-4 with a weight decay of 1e-5. The mini-batch size was set as 1. The MSPT models were trained for 150 epochs and they would early stop if the loss would not decrease in the past 30 epochs. All models were implemented by Python 3.8 with PyTorch toolkit 1.11.0 on a platform equipped with an NVIDIA GeForce RTX 3090 GPU.

### 3.4 Comparisons Experiment

**Comparison algorithms.** The proposed MSPT was compared to state-of-the-art MIL-based algorithms: 1) The traditional pooling operators, such as mean-pooling and max-pooling; 2) the attention-based algorithms, including ABMIL [8] and DSMIL [9]; 3) the Transformer-based algorithm TransMIL [11]; 4) The clustering-based algorithm ReMix [16].

**Table 1.** Comparison results on the Camelyon16 and TCGA datasets.

| Method | Camelyon16 | | TCGA-NSCLC | |
|---|---|---|---|---|
| | Accuracy | AUC | Accuracy | AUC |
| Mean-Pooling | 0.8837 | 0.8916 | 0.8911±0.011 | 0.9230±0.010 |
| Max-Pooling | 0.9147 | 0.9666 | 0.9136±0.014 | 0.9441±0.016 |
| ABMIL [8] | 0.9302 | 0.9752 | 0.9123±0.015 | 0.9457±0.017 |
| DSMIL [9] | 0.9380 | 0.9762 | 0.9049±0.010 | 0.9359±0.011 |
| TransMIL [11] | 0.9225 | 0.9734 | 0.9095±0.014 | 0.9432±0.016 |
| ReMix [16] | 0.9458 | 0.9740 | 0.9167±0.013 | 0.9509±0.016 |
| PT (Ours) | 0.9458 | 0.9809 | 0.9257±0.011 | 0.9567±0.013 |
| **MSPT (Ours)** | **0.9536** | **0.9869** | **0.9289±0.011** | **0.9622±0.015** |

**Experimental results.** Table 1 shows the comparison results on the Camelyon16 and TCGA-NSCLC datasets. In CAMELYON16, it can be found that the proposed MSPT outperforms all the compared algorithms with the best accuracy of 0.9536, and AUC of 0.9869. Compared to other algorithms, MSPT improves at least 0.78%, and 1.07% on classification ACC and AUC, indicating the effectiveness of MFFM to learn the multi-scale information of WSIs. In addition, PT achieves the best classification results in the single-resolution methods and outperforms ReMix on all indices, which proves PT can effectively re-calibrate the clustering prototypes.

In TCGA-NSCLC, the proposed MSPT algorithm again outperforms all the compared algorithms on all indices. It achieves the best classification performance of 0.9289±0.011 and 0.9622±0.015 on the ACC and AUC. Moreover, MSPT improves at least 0.78% and 1.03%, respectively, on the corresponding indices compared with all other algorithms.

### 3.5 Ablation Study

To evaluate the contribution of PT and MFFM in the proposed MSPT, we further conducted a series of ablation studies.

**Investigation of the number of prototypes in PT.** To evaluate the effectiveness of the PT, we first changed the number of prototypes $K$ in the range of $\{1, 2, 4, 8, 16, 32\}$ to get the optimal $K$ for each dataset. Then, the following two variants were compared with PT: (1) Full-bag: the first variant was only trained on all the instances; (2) Prototype-bag: the second variant was only trained on the cluster prototypes.

As shown in Fig. 3, the horizontal axes denote the number of prototypes, and the vertical axes denote the classification accuracy. In the Camelyon16 dataset, the performance of both PT and Prototype-bag increases with the increase of $K$ value, and achieves the best results with $K$=16. In the TCGA-NSCLC dataset, PT always outperforms the Full-bag and Prototype-bag. These experimental results demonstrate that PT can effectively re-calibrate the clustering prototypes to achieve superior results.

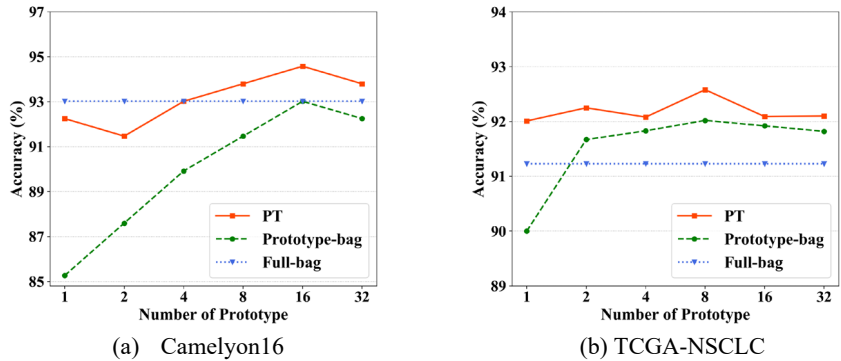

(a) Camelyon16          (b) TCGA-NSCLC

**Fig. 2.** Ablation study on the number of prototypes.

**Investigation of Multi-scale Fusion.** We further compared our MFFM with several other fusion strategies, including (1) Concatenation: this variant concatenated the cluster prototypes of each magnification before the classifier. (2) MS-Max: this variant used max-pooling on the cluster prototypes for each magnification, and then added them. (3) MS-Attention: this variant used attention-pooling [8] on the cluster prototypes for each magnification, and then added them.

Table 2 gives the results on the Camelyon16 and TCGA-NSCLC datasets. Compared with other multi-scale variants, the proposed MSPT improves ACC by at least 0.78% and 0.85% on Camelyon16 and TCGA-NSCLC, respectively, which proves that the MLP-Mixer in MFFM can effectively enhance the information communication among phenotypes and their features, thus improving the performance of feature aggregation.

**Table 2.** Classification results for evaluating different fusion strategies. All variants used the WSIs with three resolutions (5×+10×+20×).

| Method | Camelyon16 | | TCGA-NSCLC | |
|---|---|---|---|---|
| | ACC | AUC | ACC | AUC |
| Concatenation | 0.9147 | 0.9598 | 0.9147±0.018 | 0.9438±0.016 |
| MS-Max | 0.9302 | 0.9729 | 0.9203±0.014 | 0.9527±0.019 |
| MS-Attention | 0.9458 | 0.9786 | 0.9204±0.016 | 0.9571±0.012 |
| **MFFM** | **0.9536** | **0.9869** | **0.9289±0.011** | **0.9722±0.015** |

**More studies.** We provide more empirical studies, i.e., the effect of the multi-resolution scheme, the visualization results, and the training budgets, in Supplementary Materials to better understand MSPT.

## 4 Conclusion

In summary, we propose an MSPT for WSI classification that combine the prototype-based learning and multi-scale learning to generate powerful WSI-level representation. The MSPT reduces redundant instances in WSI bags by replacing instances with updatable instance prototypes, and avoids complicated procedures to align patch features at different scales. Extensive experiments validate the effectiveness of the proposed MSPT. In the future, we will develop an attention mechanism based on the magnification level to re-weight the features from different scales before fusion in MSPT.

**Acknowledgments** This work is supported by the National Natural Science Foundation of China (81871428) and 111 Project (D20031).